\documentclass[letterpaper, 10 pt, conference]{ieeeconf}  % Comment this line out if you need a4paper

\IEEEoverridecommandlockouts                              % This command is only needed if 
\usepackage[T1]{fontenc}
\usepackage{graphicx}

\usepackage{float}     % for H option in figure environment
\usepackage{hyperref}

\title{\LARGE \bf
Your Interest, Your Summaries: Query-Focused Long Video Summarization 
}

\author{Nirav Patel, Payal Prajapati, Maitrik Shah% <-this % stops a space
\thanks{Nirav Patel, Payal Prajapati, and Maitrik Shah are with the Department of Computer
Engineering, L. D. College of Engineering, Ahmedabad - 380015, Gujarat, India;
        Email:{ \{22cspnir014, payal.prajapati, maitrikshah\}@ldce.ac.in }}% <-this % stops a space
\thanks{The code is available at the \href{https://srkds.github.io/FCSNA-QFVS/}{https://srkds.github.io/FCSNA-QFVS/}} 
\thanks{© 2024 IEEE. To appear at the 18th International Conference on Control, Automation, Robotics and Vision (ICARCV), December 2024, Dubai, UAE}
}

\begin{document}

\maketitle
\thispagestyle{empty}
\pagestyle{empty}

%%%%%%%%%%%%%%%%%%%%%%%%%%%%%%%%%%%%%%%%%%%%%%%%%%%%%%%%%%%%%%%%%%%%%%%%%%%%%%%%
\begin{abstract}

Generating a concise and informative video summary from a long video is important, yet subjective due to varying scene importance. Users' ability to specify scene importance through text queries enhances the relevance of such summaries. This paper introduces an approach for query-focused video summarization, aiming to align video summaries closely with user queries. To this end, we propose the Fully Convolutional Sequence Network with Attention (FCSNA-QFVS), a novel approach designed for this task. Leveraging temporal convolutional and attention mechanisms, our model effectively extracts and highlights relevant content based on user-specified queries. Experimental validation on a benchmark dataset for query-focused video summarization demonstrates the effectiveness of our approach.

\end{abstract}

%%%%%%%%%%%%%%%%%%%%%%%%%%%%%%%%%%%%%%%%%%%%%%%%%%%%%%%%%%%%%%%%%%%%%%%%%%%%%%%%
\section{INTRODUCTION}

In recent years, there has been a significant surge of interest in video summarization, driven by the daily increase in video content creation and the growing popularity of consumption of short video formats like YouTube Shorts and Instagram Reels. The proliferation of affordable, high-quality video-capturing devices has resulted in an unprecedented volume of video data. While these videos are too long, shaky, redundant, and low-paced to watch in their entirety, they still serve as rich sources of information and knowledge. Therefore, distilling these videos to their essential content and removing redundant scenes is becoming increasingly important. As a result, video summarization, which automates this process, has gained a lot of attention in the last few years, which can produce concise summary videos, capturing the main important events without compromising the essence of the original long videos.

\begin{figure}[t]
      \centering
      \parbox{7in}{\includegraphics[width=0.5\textwidth]{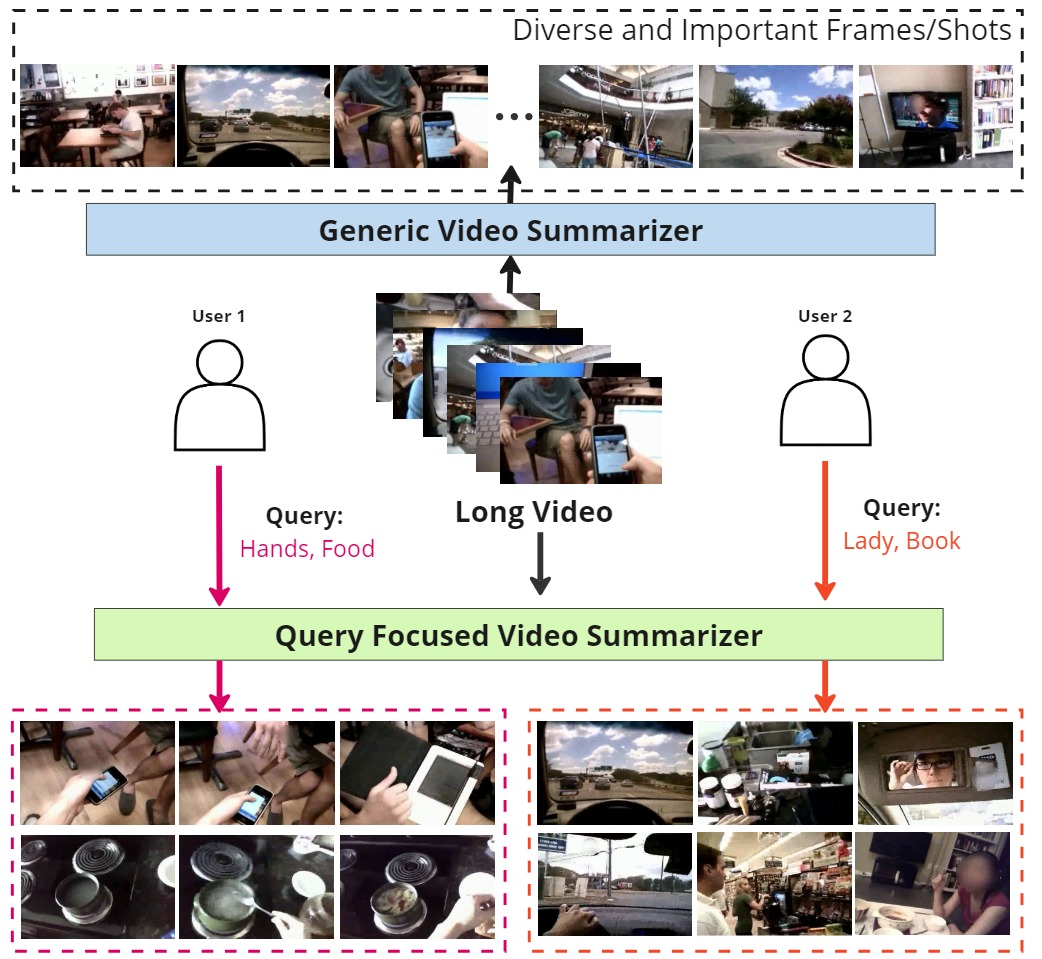}
      
}
      
      \caption{Illustration of the Difference Between Generic Video Summarizers and Query-Focused Video Summarizers for a Given Long Video and Query.}
      \label{fig1}
   \end{figure}

Previous researchers have explored generic video summarization techniques that generate concise video summaries by reducing redundant parts and selecting important and diverse shots from long videos. However, the main drawback of generic video summarization is its inability to meet the specific demands of individual users, as different users may seek different types of content from the same video shown in \autoref{fig1}. Additionally, evaluating these models poses significant challenges. This limitation has led to the development of query-focused video summarization, where the model uses user preferences specified as text queries to generate tailored video summaries. These query-focused summaries extract clips from the long video that contain the queried objects or events, along with other relevant and important scenes.

Efforts in query-focused video summarization have led to various approaches and contributions from previous authors. In \cite{sharghi_query-focused_2017} Sharghi \textit{et al.} introduced a specific dataset and evaluation matric for this task and memory-based network. Zhang \textit{et al.} in \cite{zhang_query-conditioned_2018} propose GANs with Bi-LSTMs. In \cite{xiao_convolutional_2020} Xiao \textit{et al.} propose two two-layer convolutions with an attention-based model. However, most previous methods process videos sequence by sequence, which is inefficient for long videos (3 to 5 hours). As the input video sequence grows, the computation network also becomes large for input and output and becomes a key affecting factor for the training network \cite{vaswani_attention_2017}.
Motivated by \cite{xiao_convolutional_2020}, we use 1D temporal convolution and attention mechanisms to handle long temporal relationships efficiently. Previous methods evaluate their approach using the evaluation metric proposed by Sharghi \textit{et al.} \cite{sharghi_query-focused_2017}, which measures semantic alignment using Intersection over Union (IoU) scores between machine-generated and ground truth summaries. However, we identified a limitation: without specific query regularization, high quantitative scores may occur even when selected shots don't directly match query relevance but align semantically with the ground truth. To address this, we introduced two graphs for qualitative analysis alongside our quantitative findings to provide clearer insights into the quality of our query-focused summaries.

The main contributions of this paper can be summarized as follows:
\begin{itemize}
    \item We propose the Fully Convolutional Sequence Network with Attention (FCSNA-QFVS), a novel approach for Query-Focused Video Summarization. This approach utilizes 1D temporal convolution and attention mechanisms to effectively handle long temporal relations and generate summaries in parallel.

    \item We introduce a Feature Learning Module that employs the Fully Convolution Sequence Network (FCSN) augmented with three types of attention: local self-attention, query-guided segment-level attention, and global attention. These mechanisms enable the extraction of informative shot features.

    \item We implement a shot-scoring module that takes the learned features of each shot and the query as input, predicting the query relevance score. From these scores, we select the top 2\% highest-scoring shots to generate a query-relevant video summary arranged chronologically.
\end{itemize}

For a comprehensive comparison with previous research, we evaluate our model on a benchmark UTEgocentric dataset for query-focused video summarization \cite{sharghi_query-focused_2017}, showcasing its effectiveness through both quantitative and qualitative analyses. Furthermore, we make an effort to provide a comprehensive assessment of summary quality, a dimension often overlooked in prior studies.

The remainder of this paper is structured as follows: In \autoref{sec:sec2}, we review related work on query-focused video summarization. \autoref{sec:sec3} presents our approach, providing a detailed description of each component of our model. \autoref{sec:sec4} compares our quantitative results with those of previous studies, and we conclude in \autoref{sec:sec5}.

\section{RELATED WORK}\label{sec:sec2}

In this paper, we explore the task of query-focused video summarization (QFVS).
To the best of our knowledge, Sharghi \textit{et al.} in \cite{leibe_query-focused_2016} were the first to address query-focused video summarization and  proposed a sequential hierarchical Determinantal Point Process (SH-DPP), where the first layer selects user query-relevant shots, and the second layer, conditioned on the first, identifies important shots in the video. Here, the authors relied on captions, which lacked detailed semantic information about the scene, and there was no specific evaluation metric for QFVS. To address both issues, the same authors later in \cite{sharghi_query-focused_2017} introduced a dataset and an evaluation metric specifically tailored for QFVS. This dataset has more densely annotated per-shot information. Both the dataset and the evaluation metric have emerged as a benchmark, with subsequent authors reporting their results on it.
Plummer \textit{et al.} in \cite{plummer_enhancing_2017} proposed Submod, a model that optimizes submodular objectives using vision-language embeddings. In addition, authors demonstrated that visual-language joint embeddings need not be trained on domain-specific data but can be learned from standard still image vision-language datasets and transferred to video data. In \cite{zhang_query-conditioned_2018}  Zhang \textit{et al.} introduce a GAN-based model where the generator creates video summaries, and the discriminator attempts to distinguish between ground truth and model-generated summaries. They also trained their model using a three-player loss function. Xiao \textit{et al.} in \cite{xiao_convolutional_2020} address the task by computing a shot-query similarity score, proposing a two-stage method. In the first stage, the model learns joint visual-textual feature representation, and in the second stage, it computes the shot-query relevance score. 

Fully Convolutional Sequence Networks (FCSNs) were introduced in \cite{ferrari_video_2018} for generic video summarization. Due to their state-of-the-art performance and ability to learn better visual features of long videos in parallel, we augment this architecture with attention mechanisms for the query-focused video summarization task.

Scaled dot-product attention was introduced in \cite{vaswani_attention_2017} for handling long-term dependencies in input sequences in parallel. We modify self-attention \cite{vaswani_attention_2017} to incorporate query-guided segment-level attention and global attention, which take input from both vision and text modalities.

\section{METHODOLOGY}\label{sec:sec3}

\begin{figure*}[thpb]
      \centering
      \parbox{7in}{\includegraphics[scale=0.14]{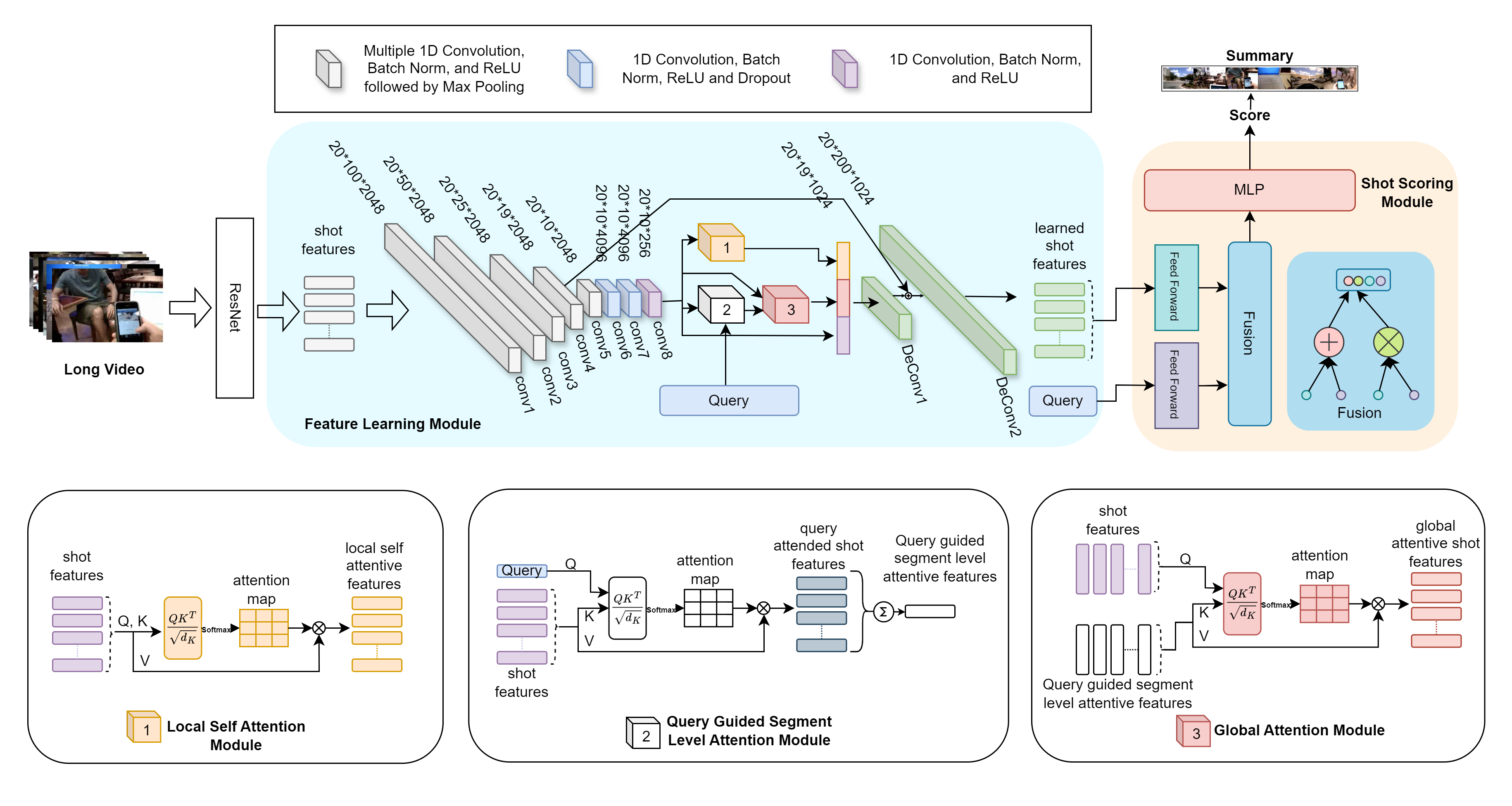}
}
      
      \caption{\textbf{Overview of FCSNA-QFVS.} Given a long video and a text query as input, we first divide the video into non-overlapping shots and group them into non-overlapping segments. Next, we pass the segmented video features to the feature learning module, where we learn visual features using eight sequential convolutional blocks. We then process these learned visual features through Local Self-Attention (LSA), Query-Guided Segment Attention (QGSA), and Global Attention (GA) to obtain locally important and globally query-guided features. We restore the original temporal length using two sequential deconvolutional layers. The feature learning network outputs the learned shot features, which we then pass to the shot scoring module to obtain a query relevance score for each shot. Finally, we generate the query-focused video summary based on these shot scores.}
      \label{fcsna}
   \end{figure*}

\subsection{PROBLEM FORMALIZATION}

Query-focused Video Summarization aims to output a video summary, containing a sequence of diverse and representative query-relevant video shots, given a long video \textbf{v} and a query \textbf{q}. To address this problem, following \cite{xiao_convolutional_2020} we formulate the task of calculating the similarity score between a shot and a query. We denote a video as a sequence of non-overlapping video shots $H_i,$ where $ i \in [1,..N]$. Each video shot refers to a fixed-length small clip of the original video. Furthermore, the video sequence is divided into nonoverlapping groups of shots called segments. We denote textual query as \textbf{$t_q$} which contains two concepts $(c_1, c_2)$. We use the lexicon of concepts created by \cite{sharghi_query-focused_2017}, each concept is a noun like 'FOOD', 'LADY', 'KIDS', etc. For each shot, we have to compute the relevance score with both concepts and then merge both scores as query-relevant scores. Finally, we can create a subset of video that is diverse and query relevant shots of the origin video based on the score. 

\subsection{FCSNA-QFVS}

The main idea of our approach is illustrated in \autoref{fcsna}. First, we propose a feature learning module to learn shot features that capture local importance relative to other shots within the segment, as well as global query relevance to shots across different segments. Then, we use a shot scoring module to assign a query relevance score to each shot’s learned feature and finally, we generate a summary based on these scores. In the following sections, we describe the feature learning module and the shot scoring module in detail.

\subsection{FEATURE LEARNING MODULE}

As shown in \autoref{fcsna}, in the feature learning module, we use 8 temporal convolution layers (1D temporal convolution block) followed by three attention mechanism blocks to handle long video relations and add textual query context, and at the end, we use two deconvolution layers(DeConv Block).  

Given a long video $V$, we extract $N$ number of non-overlapping shots. Each shot is represented as $H_i$, where $i \in [1,..., N]$. We then obtain $C$ dimensional shot feature embeddings using a pre-trained ResNet \cite{szegedy_inception-v4_2017}. We further divide these shots into $M$ non-overlapping groups of shots, referred to as segments. Each segment can contain no more than $T$ shots. These segments are denoted as $S^{T}_{i}$, where $i \in [1,..., M]$. In this context, $S_i^T$ denotes the $i$-th segment, and each segment \( S_i^T \) can contain up to \( T \) shots. The parameter \( M \) denotes the total number of segments the shots are divided into.
The final dimension of a segmented video is \( S \times T \times C \), where \( S \) is the number of segments, \( T \) is the maximum number of shots in each segment, and \( C \) is the dimensionality of the shot feature embeddings. The feature learning module takes these segmented video shot features as input and passes them to the 1D temporal convolution block. The output of the feature learning module is learned shot features.

\par
\textbf{\textit{1D Temporal Convolution Block}} To further learn the visual features of the segmented video shots, we adapt the Fully Convolutional Sequence Network (FCSN) \cite{ferrari_video_2018}, initially proposed for generic video summarization, for the task of query-focused video summarization. In 1D temporal convolution block, the input is a segmented video sequence with dimensions $1 \times S \times T \times C$, where $S$ represents the segments, $T$ is the sequence of shots in each segment, and $C$ is the feature vector of the shot. The output of 1D temporal convolution block is visual features.

As shown in \autoref{fcsna}, we retain the same eight sequential convolutional blocks from FCSN \cite{ferrari_video_2018}. The first five convolutional blocks consist of multiple temporal convolutional layers, each followed by batch normalization and ReLU activation. Temporal max-pooling is applied after each block from block 1 to block 5. Blocks 6 and 7 consist of a temporal convolutional layer followed by ReLU and dropout. Finally, block 8 includes a 1D convolutional layer followed by batch normalization and ReLU to produce the desired output channel for the subsequent attention module. The output of the 8th convolutional block is the learned temporal visual features, denoted as $C^{v}_{i}$, $i \in [1, ..., R]$, where $R$ is the reduced temporal length.

In the original FCSN \cite{ferrari_video_2018} architecture, the output of the 8th convolutional block directly feeds into the deconvolutional block. However, To incorporate query context, identify query-relevant shots, and determine locally and globally important shots within and across segments, we introduce three attention blocks between the convolutional and deconvolutional blocks:  1) local self-attention, 2) query-guided segment-level attention, and 3) global attention.

\begin{figure}[thpb]
      \centering
      \parbox{7in}{\includegraphics[width=0.5\textwidth]{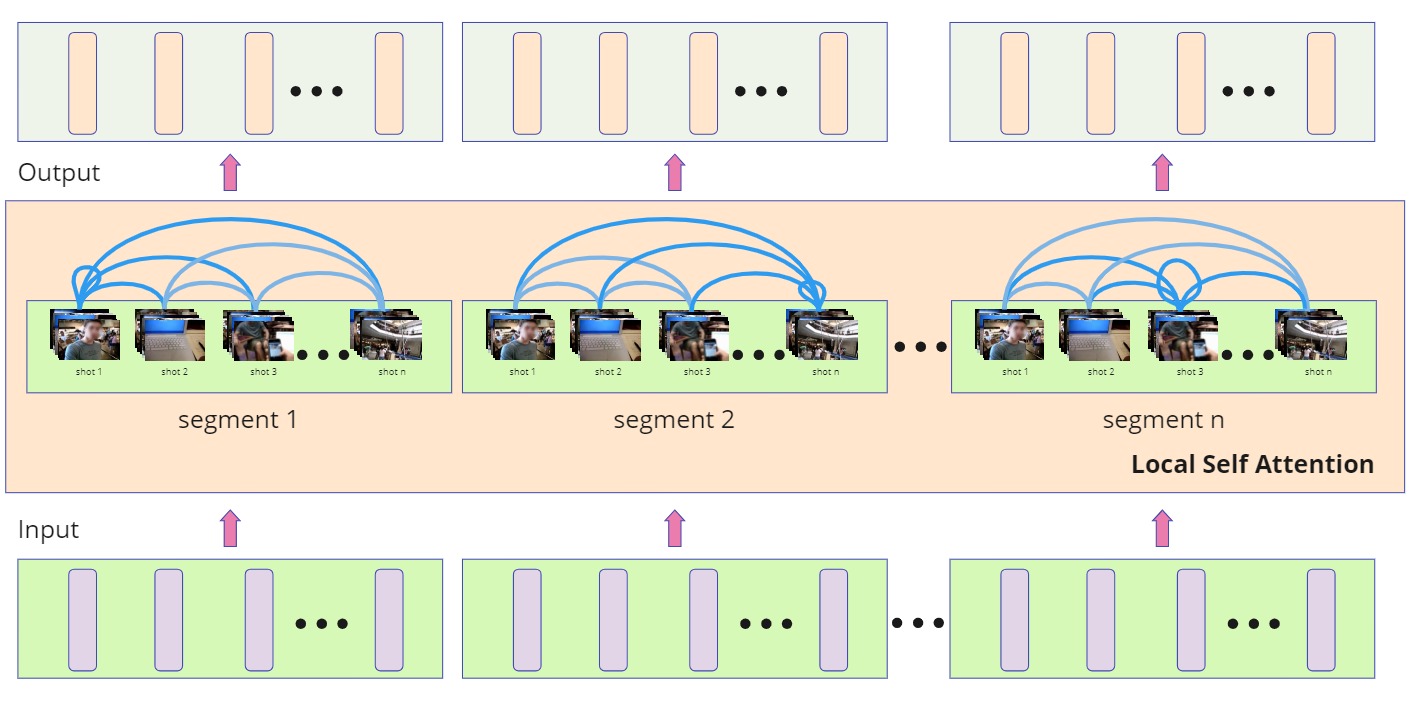}
      
}
      
      \caption{Illustration of Local Self-Attention (LSA): Finding local importance among shots within each segment for all segments.}
      \label{fig3-lsa}
   \end{figure}

% \subsection{LOCAL SELF ATTENTION}
% \subsubsection{Local Self Attetnion}
\noindent \textbf{\textit{Local Self Attention}} The Local Self-Attention (LSA) captures semantic relations among all shots within each segment and across all segments. The detailed operation is illustrated in \autoref{fcsna}, and its intuition is provided in \autoref{fig3-lsa}. To achieve this, we utilize scaled dot-product self-attention, as described in Vaswani \textit{et al.} \cite{vaswani_attention_2017}. LSA takes the learned visual shot features from the previous block as input, represented by dimensions $S \times R \times C^v$. Here, we define Query $Q_{s}$, Key $K_{s}$, and Value $V_{s}$ as follows:

\begin{center}    
$
Q_{s}, K_{s}, V_{s} = C_{i}^{v}, $ where $ \;i \in [1,...,R]
$
\end{center}
$$
C^s = Attention(Q_{s}W_{s}^{Q},K_{s}W_{s}^{K},V_{s}W_{s}^{V})
$$
$$
and \; Attn.(Q_{s},K_{s},V_{s}) = softmax(\frac{Q_{s}K_{s}^{T}}{\sqrt{d_k}})V_{s} \eqno{(1)}
$$

\noindent $W_{s}^{Q}$, $W_{s}^{K}$, and $W_{s}^{V}$ are learned parameter matrices. The output of the local self attention module is important features in local segment level denoted by $C_{i}^{s}$ where $ \;i \in [1,...,R]$ of dimention $S \times R \times C^s$

%%% query guided segment level attention
% \subsection{QUERY AWARE SEGMENT LEVEL ATTENTION}

\begin{figure}[thpb]
      \centering
      \parbox{7in}{\includegraphics[width=0.5\textwidth]{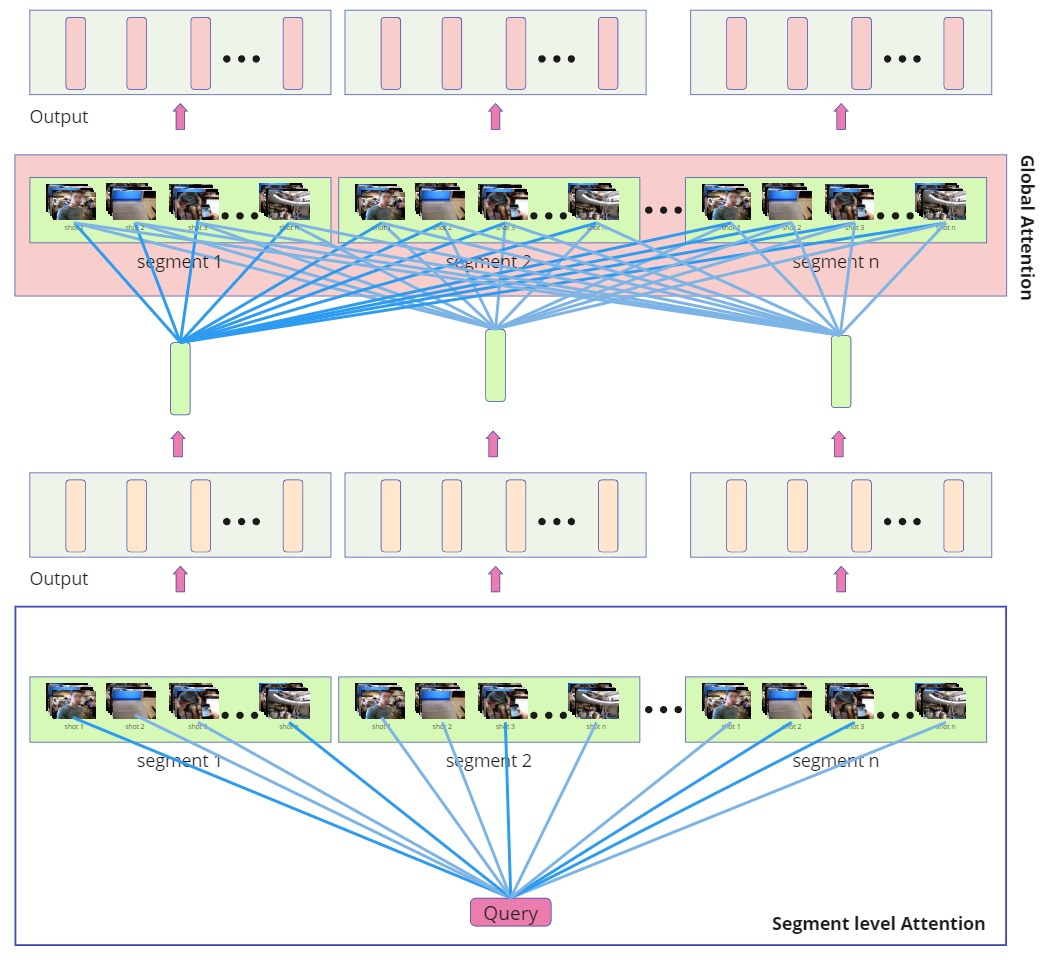}
      
}
      
      \caption{First, we extract query-guided shot features. Then, we aggregate these features to obtain query-guided segment-level attentive features. Next, we determine the global relationship among segments by evaluating the relationship between the query-guided segment-level features and each shot within a segment across all segments using global attention.}
      \label{fig4-qgsa}
   \end{figure}

\noindent \textbf{\textit{Query Guided Segment Level Attention}} To capture the semantic relation between shots and textual queries, we use Query-Guided Segment Level Attention (QGSA), as illustrated at the bottom center in \autoref{fcsna}, with its intuition depicted at the bottom of \autoref{fig4-qgsa}. The QGSA takes visual shot features $C^v$ from the 1D temporal convolution block, with dimensions $1 \times S \times R \times C^v$, and textual query features $H_q$, where $H_q$ is an average of both query concept features. The output of QGSA is a query-guided feature representation of all segments. We set Query $Q_{q}$, Key $K_{q}$, and Value $V_{q}$ as follows,

$$
Q_{q} = H_q,
$$

\begin{center}    
$
K_{q}, V_{q} = C^{v}_{i}, $ where $ \;i \in [1,...,R]
$
\end{center}

$$
C^{q} = Attention(Q_{q}W_{q}^{Q},K_{q}W_{q}^{K},V_{q}W_{q}^{V})
$$
$$
and \; Attn.(Q_{q},K_{q},V_{q}) = softmax(\frac{Q_{q}K_{q}^{T}}{\sqrt{d_k}})V_{q} \eqno{(2)}
$$

\noindent where $W_{q}^{Q}$, $W_{q}^{K}$, and $W_{q}^{V}$ are learned parameter matrices. First, QGSA generates query-guided representative features for each shot in the segment, denoted as $C^{q}_{i}$, where  $i \in [1, ..., R]$, with dimensions $1 \times S \times R \times C^q $. Here, $S$ is the number of segments, $R$ is the reduced number of temporal shots, and $C^q$ is the query-guided shot feature vector dimension. Then, QGSA aggregates each shot feature within the segment to obtain a query-guided segment-level feature representation. The final output of QGSA is a query-guided segment-level feature representation, denoted as $C^{sq}_{i}$, where $i \in [1, ..., S]$, with dimensions $1 \times S \times C^{sq}$.

%%%% Query conditioned global Attention

% \subsection{QUERY AWARE GLOBAL ATTENTION}

\noindent \textbf{\textit{Global Attention Module}} We get relations between each segment using query-guided global attention(GA) in which query-guided segment-level representative features attend each shot of the segment for all the segments. This is illustrated at the bottom right in \autoref{fcsna}, with its intuition depicted at the top of \autoref{fig4-qgsa}. The GA takes visual shot features $C^v$ from the 1D temporal convolution block, with dimensions $1 \times S \times R \times C^v$, and query-guided segment-level feature representation, denoted as $C^{sq}_{i}$, where $i \in [1, ..., S]$, with dimensions $1 \times S \times C^{sq}$ as input. We denote Query $Q_{g}$, Key $K_{g}$, and  Value $V_{g}$ as follows, 

\begin{center}   
$
Q_{g} = C_{i}^{v}, $ where $ i \; \in [1,...,R]
$
\end{center}

\begin{center}    
$
K_{g}, V_{g} = C_{j}^{sq}, $ where $ j \; \in [1,...,S]
$
\end{center}

$$
C^g = Attention(Q_{g}W_{g}^{Q},K_{g}W_{g}^{K},V_{g}W_{g}^{V})
$$
$$
and \; Attention(Q_{g},K_{g},V_{g}) = softmax(\frac{Q_{g}K_{g}^{T}}{\sqrt{d_k}})V_{g} \eqno{(3)}
$$

\noindent where $W_{g}^{Q}$, $W_{g}^{K}$, and $W_{g}^{V}$ are learned parameter matrices. We get output as global and query-guided representative features denoted as $C_{i}^{g}, \; i \in [1,..., R]$ of dimention $1 \times S \times R \times C^{g}$. 

We get the final shot feature vector by concatenating visual feature vector $C^{v}_{i}$, local self attended feature vector $C_{i}^{s}$, and global attended feature vector $C_{i}^{g}$ denoted as $C_{i}^{c} = [C^{v}_{i}; C^{s}_{i}; C_{i}^{g}] $ where $ i \in [1,...,R] $ of dimension $1 \times S \times R \times C^c$.

\noindent \textbf{\textit{DeConv Block}} As the 1D temporal convolutional block reduces the temporal length from $T$ shots to $R$ shots, we use a 1D deconvolutional block to restore the original temporal length from $R$ shots to $T$ shots. The DeconvBlock takes concated features $C^c$ as input and  outputs the learned feature vector for each shot, denoted as $C^{l}_{i}$, where $i \in [1,...,T]$ of dimension $1 \times S \times T \times C^l$.

\subsection{SHOT SCORING MODULE}

We add a shot scoring module that takes the learned visual-textual feature representation from the feature learning module and query embeddings as input, outputting a query-relevant score for each shot. First, we project the learned shot and textual query features into the same vector space using two feed-forward layers. Then, we feed both vectors into the fusion unit, which combines features by performing pointwise addition, multiplication, and concatenation of the projected feature vectors, resulting in fused feature vectors. Finally, we compute the shot query relevance score by passing these fused feature vectors through a multi-layer perceptron (MLP). From these scores, we select the top 2\% highest-scoring shots to generate
a query-relevant video summary arranged chronologically.

We train our model using binary cross-entropy loss function defined as,

$$
\mathcal{L}_{c} = \frac{1}{N}\sum_{t=1}^{N}g_{t}^{g}\log g_{t}+(1-g_{t}^{g})\log(1-g_{t}) \eqno{(4)}
$$

where $g_{t}^{g}$ is ground truth and $g_{t}$ is model predicted scores.

\section{EXPERIMENTS AND RESULTS}\label{sec:sec4}

\subsection{DATASET}

For the comparison with other work, we evaluate our model on the query-focused benchmark dataset proposed by \cite{sharghi_query-focused_2017} which is a modified version on top of the previously existing UTEgocentric \cite{lee_predicting_2015} dataset. It contains four egocentric videos, each video is 3 to 5 hours long and recorded in uncontrolled day-to-day life scenarios. The dataset provides dense per-shot annotations for supervision and to evaluate the model, where the videos are divided into small 5-second clips called shots, each labeled with several concepts present in it. In addition, the dataset contains a lexicon of 48 concise and diverse concepts, which relate to common objects in our daily lives. The dataset provides 46 queries with their ground truth summaries, each query is made up of two concepts eg \{Men, Kids\}, a total of 46 (queries) X 4 (videos). These 46 queries are made up such that they cover these four distinct scenarios: 1) all the concepts in the query appear in the same video shots together (15 such queries), 2) all concepts appear in the video but never jointly in a single shot (15 queries), 3) only one of the concepts constituting the query appears in some shots of the video (15 queries), and 4) none of the concepts in the query are present in the video (1 such query).

\subsection{PREPROCESSING}

The preprocessing steps involve two main tasks. Firstly, we extract visual features from each shot of four videos. Each video is divided into non-overlapping 5-second shots as per \cite{sharghi_query-focused_2017}. From each shot, we extract 5 frames and get 2048-dimensional visual features using a pre-trained ResNet \cite{szegedy_inception-v4_2017} model trained on Imagenet. The shot-level feature is obtained by averaging these frame features. Secondly, we divide the videos into non-overlapping segments using the KTS \cite{fleet_category-specific_2014} algorithm, ensuring a maximum of 20 segments per video, with each segment containing no more than 200 shots. For textual query features, we extract a 300-dimensional vector for each concept using publicly available GloVe \cite{pennington_glove_2014} vectors.

\subsection{EVALUATION}

We use the same evaluation metric defined in \cite{sharghi_query-focused_2017} for all our experiments and fair comparison with the other models. They make use of a bipartite graph where one side of the graph is ground truth shots and the opposite side is machine-generated shots. First, they get similarities between two ground truths and system-generated shots for that they calculate intersection over union(IoU) making use of dense per-shot annotations. For instance, if one shot is tagged by \{CAR, MEN\} and another by \{MEN, TREE, SIGN\}, then the IOU similarity between them is 1/4 = 0.25, and these similarity scores between two shots become the weight of the edge. Afterward, calculate the maximum weight matching of a bipartite graph, then calculate the sum of the weight of edges of matched pairs. Finally, divide the sum of weight by the total length of the machine summary to get precision, divide the sum of weight by the total length of ground truth summary to get recall, and with precision and recall we can get F1 as harmonic meaning of both.

\subsection{IMPLEMENTATION DETAILS}

We use PyTorch to implement our method. We reduce the temporal length of a segment from 200 to 10 using blocks 1 to 8 and set its output feature dimension to 256. We set a dropout rate of 0.3 in blocks 6 and 7. For all three single-head self-attention, query-guided segment-level attention, and global attention we set 256 as head size. We set 1024 channel dimensions for the output of the DeConv block. In both linear projection layers for learned visual features and textual features, the output dimension is set to 300. We performed 4 experiments by selecting one by one 1 video for testing and the rest for training. We use Adam optimizer \cite{kingma_adam_2014} with a learning rate 0.0001 and a decay rate of 0.8. We train our model for 20 epochs with a mini-batch size of 5 which takes about 1 hour on a single Nvidia T4 16 GB GPU card.

\subsection{QUANTITATIVE RESULTS}

\begin{table*}[h]
    \centering
    \caption{Comparison results on the query-focused video summarization dataset in terms of Precision, Recall and F1-score.}
    \resizebox{\textwidth}{!}{% for table to fit to page
\begin{tabular}[b]{c|c|c|c|c|c|c|c|c|c|c|c|c|c|c|c}
        \hline
         & \multicolumn{3}{c|}{SH-DPP \cite{leibe_query-focused_2016}} & \multicolumn{3}{c|}{QC-DPP \cite{sharghi_query-focused_2017}} & \multicolumn{3}{c|}{ TPAN \cite{zhang_query-conditioned_2018}} & \multicolumn{3}{c|}{CHAN \cite{xiao_convolutional_2020}}  & \multicolumn{3}{c}{Ours} \\
        \hline
         & Pre & Rec & F1 & Pre & Rec & F1 & Pre & Rec & F1 & Pre & Rec & F1 & Pre & Rec & F1 \\
         \hline
        Video 1 & 50.56 & 29.64 & 35.67 & 49.86 & 53.38 & 48.68 & 49.66 & 50.91 & 48.74 & 54.73 & 46.57 & \textbf{49.14} & 55.89 & 39.41 & 45.15\\
        \hline
        Video 2 & 42.13 & 46.81 & 42.72 & 33.71 & 62.09 & 41.66 & 43.02 & 48.73 & 45.30 & 45.92 & 50.26 & 46.53 & 47.47 & 54.38 & \textbf{50.32}\\
        \hline
        Video 3 & 51.92 & 29.24 & 36.51 & 55.16 & 62.40 & 56.47 & 58.73 & 56.49 & 56.51 & 59.75 & 64.53 & \textbf{58.65} & 67.88 & 49.93 & 57.24  \\
        \hline
        Video 4 & 11.51 & 62.88 & 18.62 & 21.39 & 63.12 & 29.96 & 36.70 & 35.96 & 33.64 & 25.23 & 51.16 & 33.42 & 28.22 & 56.40 & \textbf{37.20}\\
        \hline

        AVG & 39.03 & 42.14 & 33.38 & 40.03 & 60.25 & 44.19 & 47.03 & 48.02 & 46.05 & 46.40 & 53.13 & 46.94 & \textbf{49.86} & 50.03 & \textbf{47.47}\\
        \hline
\end{tabular}%
    }
    
    \label{tab:restab}
\end{table*}

\begin{figure*}[thpb]
      \centering
      \parbox{7in}{\includegraphics[scale=0.31]{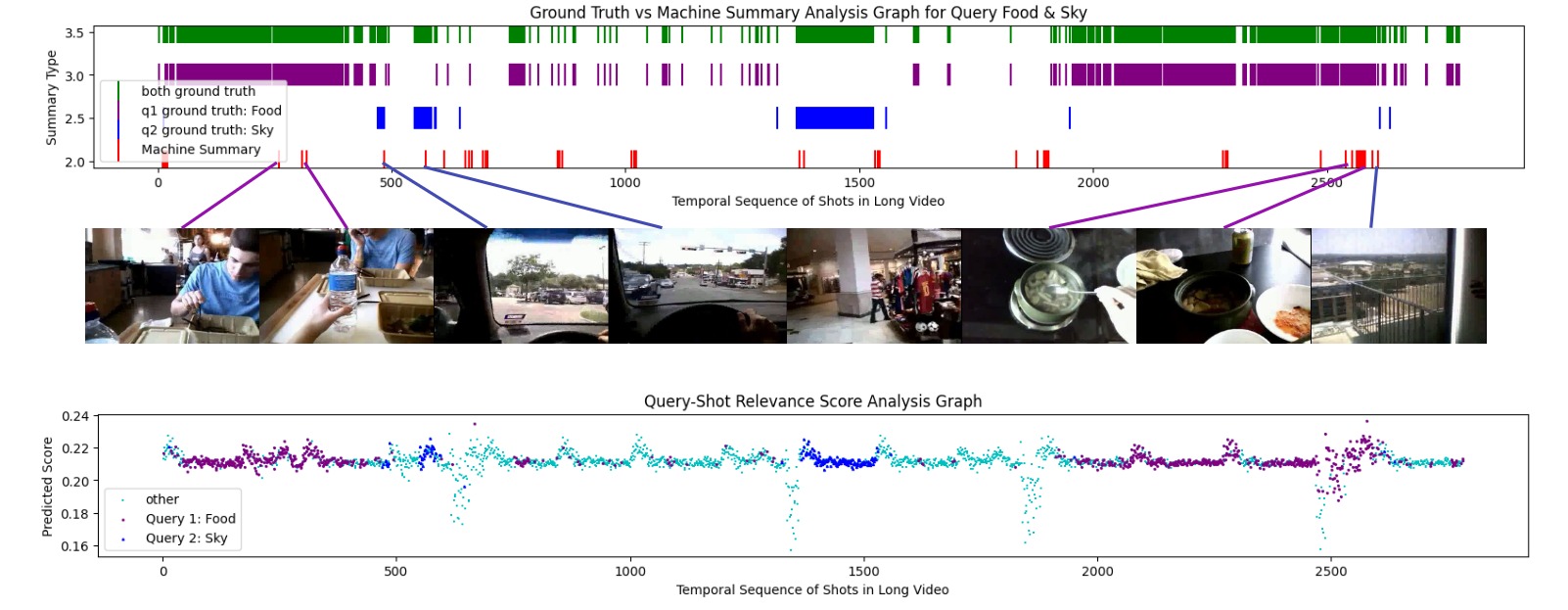}
}
      
      \caption{Illustration of our qualitative results for queries 'Food' and 'Sky'}
      \label{fig5-qr}
   \end{figure*}
   
In \autoref{tab:restab}, we compare our approach's precision, recall, and F1 scores with SH-DPP \cite{leibe_query-focused_2016}, QC-DPP \cite{sharghi_query-focused_2017}, TPAN \cite{zhang_query-conditioned_2018}, and CHAN \cite{xiao_convolutional_2020}. Following \cite{sharghi_query-focused_2017}, we conducted four experiments, selecting one video for testing and the remaining for training. Our approach achieves an average F1 score of 47.47\%, outperforming the baseline (46.94\%) by 1.12\%. Specifically, our approach achieves a 50.32\% F1 score for video 2, surpassing the baseline (46.53\%) by 11.31\%, and records a 37.20\% F1 score for video 4, outperforming the baseline (33.42\%) by 8.14\%.

\subsection{QUALITATIVE RESULTS}

In this section, we propose two qualitative analysis graphs inspired by \cite{zhang_query-conditioned_2018} and \cite{xiao_convolutional_2020}, addressing the following questions: 1) From which part of the long video does the model select shots? 2) Are these shots query-relevant or irrelevant? 3) How many shots in the summary are query-relevant versus irrelevant? Lastly, 4) what scores does the model assign to query-relevant and irrelevant shots?

To address first three questions, we propose a ground truth vs machine summary analysis graph, depicted at the top of \autoref{fig5-qr}. Previous studies \cite{zhang_query-conditioned_2018} and \cite{xiao_convolutional_2020} typically relied on ground truth summary shots provided in dataset [1]. In contrast, we utilize dense per-shot annotations that comprehensively cover the entire temporal length for a given query. The x-axis represents the sequence of shot appearances, while the y-axis illustrates different types of summaries (i.e., ground truth and machine-generated summary). We distinguish ground truth summaries for each query with blue and purple colors, showing their union in green. Additionally, we highlight the top 2\% selected shots from the model-generated summary in red.

To address the fourth question, we propose a query-shot relevance score analysis graph shown at the bottom of \autoref{fig5-qr}. As users interact with the model using textual queries, understanding how the model adjusts its predictions to changes in user queries becomes crucial. This graph facilitates the interpretation of query relevance scores (ranging from 0 to 1) assigned to each shot throughout the video, leveraging dense-per-shot annotations as defined in \cite{sharghi_query-focused_2017}. 

The x-axis represents a sequence of shots, where each point corresponds to a shot in the entire timeline. The y-axis at each x-axis point represents the model-predicted relevance score (between 0 and 1). Shot interpretations are color-coded as follows: shots related to query one are marked purple, shots related to query two are marked blue, and shots not relevant to any query are marked cyan.

In Figure 5, we present our qualitative results for the queries 'Food' and 'Sky'. Our approach generates a summary comprising 22 shots of scenes containing food and 3 shots of scenes featuring the sky, out of a total of 55 selected shots. Additionally, the graph at the bottom of Figure 5 illustrates that our method assigns high scores to query-relevant shots, along with other important shots from the long video.

\section{CONCLUSION}\label{sec:sec5}

We have introduced a Fully Convolutional Sequence Network with Attention (FCSNA-QFVS) for query-focused video summarization. By adapting a popular generic video summarization network FCSN and enhancing it with an attention mechanism and shot scoring module, our approach effectively captures query relevance and long-distance relationships in parallel, overcoming the limitations of sequence networks like LSTMs. We demonstrate the quantitative and qualitative effectiveness of our approach through extensive experiments on a benchmark dataset of long videos specifically created for query-focused video summarization.

%%%%%%%%%%%%%%%%%%%%%%%%%%%%%%%%%%%%%%%%%%%%%%%%%%%%%%%%%%%%%%%%%%%%%%%%%%%%%%%%

\section*{ACKNOWLEDGMENT}

We thank the Computer Engineering Department at L. D. College of Engineering, Ahmedabad, for providing access to NVIDIA GPUs, which were used extensively for conducting our experiments.

\bibliographystyle{IEEEtran}
\UseRawInputEncoding\bibliography{ref}

\end{document}